\documentclass[12pt]{article}
\usepackage{amsfonts}
\usepackage[utf8]{inputenc}
\usepackage{hyperref}
\usepackage{amssymb,amsmath,array}
\usepackage{caption}
\usepackage{subcaption}

 \usepackage[pdftex]{graphicx}
 \graphicspath{AggInter}
 \graphicspath{TC_V2}
  \DeclareGraphicsExtensions{.pdf}

\begin{document}

%--------------------------------------------------------------------------

\begin{center}
{\Large
	{\sc  Modelling time evolving interactions in networks through a non stationary extension of stochastic block models}
}
\bigskip

 Marco Corneli, Pierre Latouche and Fabrice Rossi
 
\bigskip

{\it
 Universit\'e Paris 1 Panth\'eon-Sorbonne - Laboratoire SAMM \\
90 rue de Tolbiac, F-75634 Paris Cedex 13 - France
 
}
\end{center}
\bigskip

%--------------------------------------------------------------------------

{\bf R\'esum\'e.}
Le modèle à blocs stochastiques (SBM) décrit les interactions entre les sommets d'un graphe selon une approche probabiliste, basée sur des classes latentes. SBM fait l'hypothèse implicite que le graphe est stationnaire. Par conséquent, les interactions entre deux classes sont supposées avoir  la même intensité pendant toute la période d'activité. Pour relaxer l'hypothèse de stationnarité, nous proposons une partition de l'horizon temporel en sous intervalles disjoints, chacun de même longueur. Ensuite, nous proposons une extension de SBM qui nous permet de classer en même temps les sommets du graphe et les intervalles de temps où les interactions ont lieu. Le nombre de classes latentes (K pour les sommets, D pour les intervalles de temps) est enfin obtenu à travers la maximisation de la vraisemblance intégrée des données complétées (ICL exacte). Après avoir testé le modèle sur des données simulées, nous traitons un cas réel. Pendant une journée, les interactions parmi les participants de la conférence HCM Hypertext (Turin, 29 Juin – 1er Juillet 2009) ont été traitées. Notre méthodologie nous a permis d'obtenir une classifications intéressante des 24 heures: les moments de rencontre tels que les pauses café ou buffets ont bien été détectés. La complexité de l'algorithme de recherche, linéaire en fonction du  nombre initial de clusters ($K_{max}$ et $D_{max}$ respectivement), nous oriente vers l'utilisation d'instruments avancés de classification, pour réduire le nombre attendu de classes latentes et ainsi pouvoir utiliser le modèle pour des réseaux de grand dimension.                     
\smallskip

{\bf Mots-cl\'es.} Graphes al\'eatoires, classification temporelle, mod\`eles \`a blocs stochastiques, vraisemblance classifiante int\'egr\'ee 
\bigskip\bigskip

{\bf Abstract.} 
In this paper, we focus on the stochastic block model (SBM),
a probabilistic tool describing interactions between nodes of a network 
using latent clusters. The SBM assumes that the network
has a stationary structure, in which connections of 
time varying intensity are not taken into account. 
In other words, interactions between two groups are forced to have the same 
features during the whole observation time. To overcome this limitation,
we propose a partition of the whole time horizon, in which interactions are observed, 
and develop a non stationary extension of the SBM,
allowing to simultaneously cluster the nodes in a network along with 
fixed time intervals in which the interactions take place. The number 
of clusters (K for nodes, D for time intervals) as well as the class memberships are finally
obtained through maximizing the complete-data integrated likelihood 
by means of a greedy search approach. After showing  that the model 
works properly with simulated data, we focus on a real data set.  
We thus consider the three days \emph{ACM Hypertext} conference held in Turin,
June 29th - July 1st 2009. Proximity interactions between attendees 
during the first day are modelled and an interesting
clustering of the daily hours is finally obtained, with times of 
social gathering (e.g. coffee breaks) recovered by the approach. 
Applications to large networks are limited due to the computational 
complexity of the  greedy search which is dominated by
the number $K_{max}$ and $D_{max}$ of clusters used in the initialization. Therefore,
advanced clustering tools are considered to reduce the number of 
clusters expected in the data, making the greedy search applicable to 
large networks.

\smallskip


%--------------------------------------------------------------------------

\section{Introduction}

\begin{figure}[!h]
  \centering
  \includegraphics[width=0.5\textwidth]{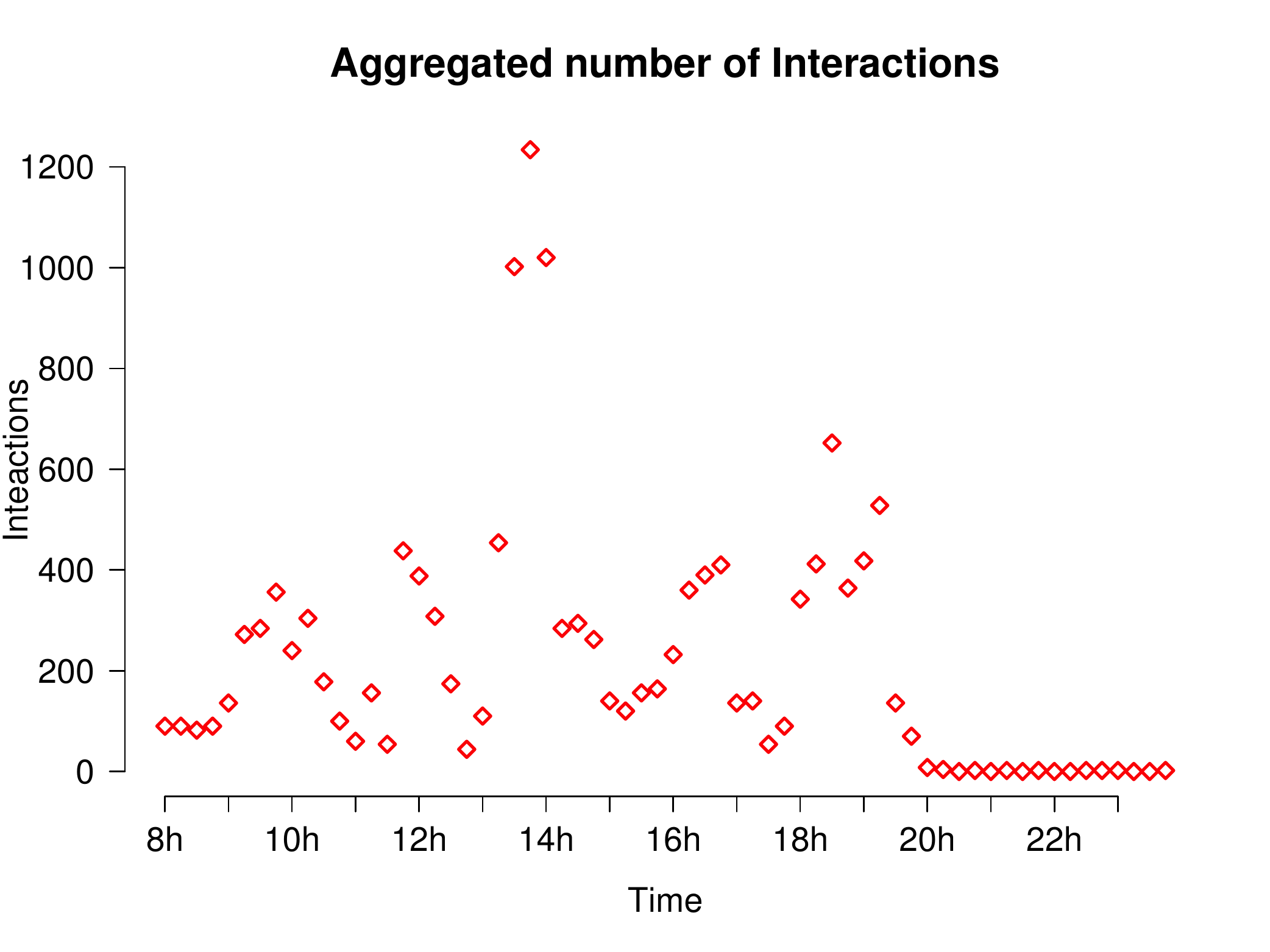}
  \caption{\footnotesize Aggregated number of interactions between conference attendees per quarter-hour.}
  \label{fig:Agginter_prima}
\end{figure}

Since the interactions between nodes of a network generally have a time varying intensity, the network has a non trivial time structure that we wish to infer. An example of this complexity can be observed in Figure \eqref{fig:Agginter_prima}. On the vertical axis the aggregated number of proximity face-to-face interactions (less than 1.5 meter) between attendees of the \emph{HCM Hypertext} conference (Turin, June 29th - July 1st, 2009) is given. On the horizontal axis, a time line is reported, corresponding to the 24 hours of the first day of conference. 
The day was partitioned in small time intervals of 20 seconds in the original data frame.  We considered 15 minutes time aggregations leading to a partition of the day made of 96 consecutive quarter-hours: each red point in the figure corresponds to one of them. If we associate each attendee to a node and each interaction to an edge, the non stationarity of such a graph is clear. In the following section, after describing the general stochastic block model, we illustrate the temporal evolution we propose.

\section{A non stationary stochastic block model}
%\subsection{Latent Block model}
We present here the SBM (Holland et al. \cite{Holland}), using the same notations as in Wyse, Friel, Latouche (2014). The set of nodes $A=\{a_1, \dots, a_{N} \}$ is introduced.  Undirected links between nodes $i$  and $j$ from $A$ are counted by the observed variable $X_{ij}$, being the component $(i,j)$ of the $N \times N$ adjacency matrix $X=\{X_{ij}\}_{i \leq N, j \leq N}$. Nodes in $A$ are clustered in $K$ disjointed subgroups respectively: 
\begin{equation*}
 A= \cup_{k \leq K}A_k, \qquad A_l \cap A_g=\emptyset, \quad\forall l\neq g  
\end{equation*}
Nodes in the same cluster in $A$ have linking attributes of the same nature. 
We introduce an hidden vector $\mathbf{c}=\{c_1,\dots, c_N\}$ labelling each node's membership:
\begin{equation*}
 c_i=k\qquad\textit{iff}\quad{a_i \in A_k,\quad \forall k\leq K}.
\end{equation*}

%\subsection{A non  stationary approach}
In order to introduce a temporal dimension, consider now a sequence of equally spaced, adjacent time steps $\{\Delta_u:= t_u - t_{u-1} \}_{u \leq U}$ over the interval $[0,T]$ and a partition $C_1, \dots, C_D$ of the same interval\footnote{$T$ and $U$ are linked by the following relation: $T=\Delta_u U$.}. We introduce furthermore a random vector $\mathbf{y}=\{y_u\}_{u \leq U}$, such that $y_u=d$ if and only if $I_u:=]t_{u-1}, t_{u}] \in C_d, \forall d \leq D$. 
We attach to $\mathbf{y}$ a multinomial distribution:
\begin{equation*}
 p(\mathbf{y} | \boldsymbol{\beta},D)=\prod_{d \leq D}\beta_d^{|C_d|},
\end{equation*}
where $|C_d|=\#\{I_u : I_u \in C_d\}$. Now we define $N^{I_u}_{ij}$ as the number of observed connections between $a_i$ and $a_j$, in the time interval $I_u$ and we make the following crucial assumption:
\begin{equation}
 p(N^{I_u}_{ij}| c_i=k, c_j=g, y_u=d)  \qquad\text{follows a}\qquad\text{Poisson}(\Delta_u\lambda_{kgd}),
\end{equation}
hence the number of interactions is \emph{conditionally} distributed like a Poisson random variable with parameter depending on $k,g,d$ ($\Delta_u$ is constant).

% But the parameter of this distribution varies now as a function of the membership of $I_u$, in other words if we note: 
% \begin{equation*}
%   Y_u:=N^{I_u}_{ij}| c_i=k, w_j=g,
% \end{equation*}
% the variables $Y_1, \dots, Y_U$ are no longer identical distributed.

\medskip
\textbf{Notation}: In the following, for seek of simplicity, we will note:
\begin{equation*}
\prod_{k,g,d} := \prod_{k \leq K} \prod_{g \leq K}\prod_{ d \leq D} \quad\text{and}\quad \prod_{c_i}:= \prod_{i:c_i=k}
\end{equation*}
and similarly for $\prod_{c_j}$ and $\prod_{y_u}$.

\medskip
The adjacency matrix, noted $N^{\Delta}$, has three dimensions ($N\times N \times U$) and its observed likelihood can be computed explicitly:
\begin{equation}
\label{eq:obsLL}
 p(N^{\Delta}|\Lambda, \mathbf{c}, \mathbf{y}, K,D)= \prod_{k,g,d} \frac{\Delta^{S_{kgd}}}{\prod_{c_i} \prod_{c_j}\prod_{y_u} N_{ij}^{I_u}! }e^{-\Delta \lambda_{kgd} R_{kgd}}\lambda_{kgd}^{S_{kgd}}, 
 \end{equation}
where we noted $S_{kgd}:= \sum_{c_i}\sum_{c_j} \sum_{y_u} N^{I_u}_{ij}$ and $R_{kgd}:=|A_k||A_g||C_d|$\footnote{Self loops are considered here for seek of simplicity. The model can easily be extended to graphs with undirected links and/or no self loops.}. The subscript $u$ was removed from $\Delta_u$ to emphasize that time steps are equally spaced for every $u$.

Since $\mathbf{c}$ and $\mathbf{y}$ are not known, a multinomial factorizing probability density $p(\mathbf{c}, \mathbf{y}| \Phi, K,D)$, depending on a hyperparameter $\Phi$, is introduced. The joint distribution of labels looks finally as follows:
 \begin{equation}
 \label{eq:term_1_ns}
   p(\mathbf{c}, \mathbf{y} |\Phi, K, D) = \left(\prod_{k \leq K}\omega_k^{|A_k|}\right)\left(\prod_{d \leq D}\beta_d^{|C_d|}\right),
\end{equation}
where $\Phi=\{\boldsymbol{\omega},\boldsymbol{\beta} \}$.

% To model such a dynamic, we introduce a non homogeneous Poisson process counting interactions between hidden clusters. Thus, for a fixed time interval $I_t$ and a couple of nodes $(i,j)$, we introduce a Poisson process counting interactions between $i$ and $j$, whose intensity varies as a function of  both the latent class the two nodes belong to and the latent class the interval $I_t$ belongs to.
% The class memberships are modelled using multinomial distributions and, following a Bayesian perspective, we attached ad hoc \emph{a priori} distributions to the model parameters to obtain an analytical expression of the integrated likelihood of complete-data (exact ICL). To estimate the number of clusters (K for nodes, D for time intervals) as well as memberships to the different classes, a greedy search approach is used to numerically maximize the ICL. We illustrate our results and make a comparison with those obtained by using other models. We focus on the advantages of the approach proposed with respect to a standard SBM.  
\subsection{Exact ICL for non stationary SBM}
The integrated classification criterion (ICL) was introduced as a model selection criterion in the context of Gaussian mixture models by Biernacky et al. \cite{Biernacky}. C\^{o}me and Latouche \cite{Latouche1} proposed an exact version of the ICL based on a Bayesian approach for the stochastic block model and this is the approach we follow here. 
The quantity we focus on is the \emph{complete data} log-likelihood, integrated with respect to the model parameters $\Phi$ and $\Lambda=\{\lambda_{kgd}\}_{k\leq K, g \leq K, d\leq D}$:
\begin{equation}
 \mathcal{ICL}= \log\left( \int p(N^{\Delta},\mathbf{c}, \mathbf{y}, \Lambda, \Phi| K,D)d\Lambda d\Phi \right).
\end{equation}
Introducing a prior distribution $\nu(\Lambda, \Phi | K,D)$ over the pair $\Phi, \Lambda$ and thanks to ad hoc independence assumptions,
% \begin{equation*}
%    \nu(\Lambda, \boldsymbol{\omega}, \boldsymbol{\rho}, \boldsymbol{\beta}| K,D)=\nu(\Lambda|K,G,D) \nu(\boldsymbol{\omega}|K) \nu(\boldsymbol{\rho} | G)\nu(\boldsymbol{\beta}|D),
% \end{equation*}
the ICL can be rewritten as follows:
\begin{equation}
\label{eq: ICLNI}
 \mathcal{ICL}= \log\left(\nu(N^{\Delta}| \mathbf{c}, \mathbf{y},K,D)\right) + \log\left( \nu(\mathbf{c}, \mathbf{y}| K,D)\right).
\end{equation}
The choice of prior distributions over the model parameters is crucial to have an explicit form of the ICL.

\subsection{A priori distributions}
We consider the conjugate prior distributions. Thus we impose a Gamma a priori over $\Lambda$:
\begin{equation*}
  \nu(\lambda_{kgd}| a, b)=\frac{b^a}{\Gamma(a)}\lambda_{kgd}^{a-1}e^{-b\lambda_{kgd}}
\end{equation*}
and a factorizing Dirichlet a priori distribution to $\Phi$:
\begin{equation*}
 \nu(\Phi|K,D) = \text{Dir}_K(\boldsymbol{\omega}; \alpha,\dots,\alpha) \times  \text{Dir}_D(\boldsymbol{\beta}; \gamma,\dots,\gamma).
\end{equation*}
It can be proven  that the two terms in \eqref{eq: ICLNI}, reduce to:
\begin{align}
 \label{eq:ICLI1}
   \nu(N^{\Delta}| \mathbf{c}, \mathbf{y},K,D) =&  \prod_{k,g,d} \frac{b^a\Delta^{S_{kgd}}}{\Gamma(a) \prod_{c_i} \prod_{c_j}\prod_{y_u} N_{ij}^{I_u}! }\\
   &\frac{\Gamma(S_{kgd}+a)}{[\Delta R_{kgd}+ b]^{S_{kgd} + a}} \nonumber
\end{align}
and:
\begin{align}
 \nu(\mathbf{c}, \mathbf{y}| K,D)=\frac{\Gamma(\alpha K)}{\Gamma(\alpha)^K}\frac{\prod_{k\leq K}\Gamma(|A_k| + \alpha)}{ \Gamma(N + \alpha K)}
 \times \frac{\Gamma(\gamma D)}{\Gamma(\gamma)^D}\frac{\prod_{d\leq D}\Gamma(|C_d| + \gamma)}{ \Gamma(U + \gamma D)}.
 \end{align}

\section{ICL maximization and experiments}
In order to maximize the integrated complete likelihood (ICL) in equation \eqref{eq: ICLNI} with respect to the four unknowns $\mathbf{c}, \mathbf{y}, K, D$, we rely on a greedy search over labels and the number of nodes and time clusters. This approach is described in Wyse, Frial and Latouche \cite{Latouche2} for a stationary latent block model. 

\section{Experiments}
\subsection{Simulated data}

Some experiments on simulated data have initially been conducted. Based on the model described above, we simulated interactions between 50 nodes, clustered in three groups ($K=3$). Interactions take place into 24 time intervals of unitary length (ideally one hour), clustered into three groups too ($D=3$). 
% \begin{center}
%  \begin{tabular}{|c|c|}
%   \hline
%  \multicolumn{2}{|c|}{Parameters} \\
%   \hline
%  N & 50 \\
%  \hline
%  M & 50 \\
%  \hline
%  U & 24 \\
%  \hline
%  K & 3 \\
%  \hline
%  G & 3 \\
%  \hline 
%  D & 3 \\
%  \hline 
%  $\Delta$ & 1\\
%  \hline
%  \end{tabular}
% \end{center}
Nodes and time intervals labels are sampled from multinomial distributions. 
With these settings, we consider 27 different Poisson parameters ($\lambda$s) generating connections between nodes. The generative model used to produce them is described by:

\begin{equation*}
  \lambda_{kgl}= s_1[k] + s_2[g] + s_3[l], \qquad\ k,g,l \in \{1, 2, 3\}
\end{equation*}
where:

\begin{equation*}
 s_1=[0,2,4] \quad s_2=[0.5,1,1.5] \quad s_3=[0.5,1,1.5]
\end{equation*}
ans $s_1[k]$ denotes the k-th component of $s_1$. Similarly for $s_2$ and $s_3$. 
The greedy search algorithm we coded was able to exactly recover these initial settings, converging to the true ICL of $-122410$. Other experiments were run with different values inside vectors $s_1, s_2, s_3$. Not surprisingly  the more nuanced differences between $\lambda$s are, the more difficult it is for the algorithm to converge to the true value of the ICL\footnote{Greedy search algorithms are path dependent and they could converge to local maxima.}.

\subsection{Real Data}

The dataset we used was collected during the \textbf{ACM Hypertext} conference held in Turin, June 29th - July 1st, 2009. Conference attendees volunteered to to wear radio badges that monitored their face-to-face proximity. The dataset represents the dynamical network of face-to-face proximity of 113 conference attendees over about 2.5 days\footnote{More informations can be found at: 

\url{http://www.sociopatterns.org/datasets/hypertext-2009-dynamic-contact-network/ }.

}. Further details can be found in Isella, Stehl\'e, Barrat, Cattuto, Pinton, Van den Broeck \cite{Cattuto}. We focused on the first conference day, namely the twenty four hours going from 8am of June 29th to 7.59am of June 30th. The day was partitioned in small time intervals of 20 seconds in the original data frame.  We considered 15 minutes time aggregations, thus leading to a partition of the day made of 96 consecutive quarter-hours ($U=96$ with previous notation). A typical row of the adjacency matrix we analyzed, looks like:

\begin{center}
 \begin{tabular}{c|c|c|c}
  \hline
  \footnotesize\emph{Person 1} & \footnotesize\emph{Person 2} & \footnotesize\emph{Time Interval (15m)} & \footnotesize\emph{Number of interactions} \\
  \hline
   52 & 26 & 5 & 16 \\
  \hline 
 \end{tabular}
\end{center}
It means that conference attendees 52 and 26, between 9am and 9.15am have spoken for $16 \times 20s \approx 5m30s$.  

The greedy search algorithm converged to a final ICL of -53217.4, corresponding to 23 clusters for nodes (people) and 3 time clusters. 
\begin{figure}[!h]
\centering
\begin{subfigure}{.5\textwidth}
  \centering
  \includegraphics[width=\linewidth]{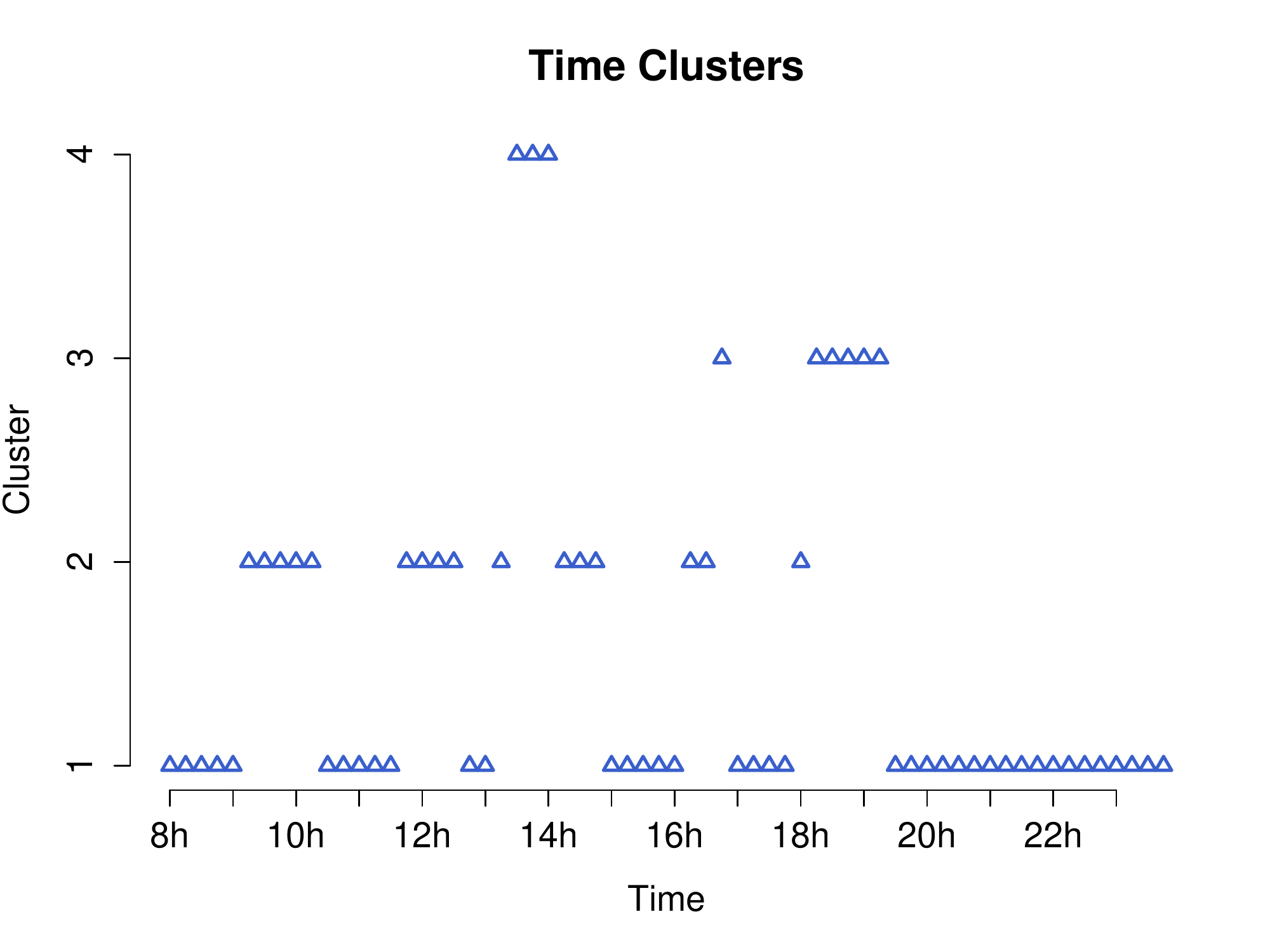}
  \caption{\footnotesize Clustered time intervals.}
  \label{fig:TC}
\end{subfigure}%
\begin{subfigure}{.5\textwidth}
  \centering
  \includegraphics[width=\linewidth]{AggInter}
  \caption{\footnotesize Connections for every time interval.}
  \label{fig:Agginter}
\end{subfigure}
\caption{\small The aggregated connections for every time interval (\ref{fig:Agginter}) and time clusters found by our model (\ref{fig:TC}) are compared.}
\label{fig:MainFIgure}
\end{figure}
In Figure \eqref{fig:TC} we show how daily quarter-hours are assigned to each cluster: the class $C_1$ contains intervals marked by a weaker intensity of interactions (on average), whereas  intervals inside $C_3$ are characterized by the highest intensity of interactions. This can either be verified analytically by averaging estimated Poisson intensities for each one of the three clusters or graphically by looking at Figure \eqref{fig:Agginter}. In this Figure we computed the total number of interactions between conference attendees for each quarter-hour and it can clearly be seen how time intervals corresponding to the higher number of interactions have been placed in cluster $C_3$, those corresponding to an intermediate interaction intensity, in $C_2$ and so on. It is interesting to remark how the model can quite closely recover times of social gathering like the lunch break (13.00-15.00) or the ``wine and cheese reception'' (18.00-19.00). 
% \begin{itemize}
%  \item 9.00-10.30 - set-up time for posters and demos.
%  \item 13.00-15.00 - lunch break.
%  \item 18.00-19.00 - wine and cheese reception.
% \end{itemize}
A complete program of the day can be found at:

\url{http://www.ht2009.org/program.php}.

\section{Conclusions}

We proposed a non-stationary evolution of the latent block model (LBM) allowing us to simultaneously infer the time structure of a bipartite network  and cluster the two node sets. The approach we chose consists in partitioning the entire time horizon in fixed-length time intervals to be clustered on the basis of the intensity of connections in each interval. We derived the complete ICL for such a model and maximized it numerically, by means of a greedy search, for two different networks: a simulated and a real one. The results of these two tests highlight the capacity of the model to capture non-stationary time structures.

% 
% \section*{Bibliographie}
% \noindent [1] Auteurs (ann\'ee), Titre, revue, localisation.
% 
% \noindent [2] Achin, M. et Quidont, C. (2000), {\it Th\'eorie des
% Catalogues}, Editions du Soleil, Montpellier.
% 
% \noindent [3] Noteur, U. N. (2003), Sur l'int\'er\^et des
% r\'esum\'es, {\it Revue des Organisateurs de Congr\`es}, 34, 67--89.


\begin{thebibliography}{99}

\bibitem{Govaert} A.N. Randriamanamihaga, E. C\^ome, L. Oukhellou and G. Govaert, Clustering the V\'elib' dynamic Origin/Destination flows using a family of 
Poisson mixture models, \emph{Neurocomputing}, vol. 141, pp. 124-138, 2014.

\bibitem{Holland} P.W.Holland, K.B. Laskey and S. Leinhardt, Stochastic blockmodels: first steps, \emph{Social networks}, vol.5, pp.109-137, 1983.  

\bibitem{Rossi} R. Guigour\`es, M. Boull\'e and F. Rossi, A Triclustering Approach for Time Evolving Graphs, 
in \emph{Co-clustering and Applications, IEEE 12th International Conference on Data Mining Workshops (ICDMW 2012)},
pages 115-122, Brussels, Belgium, December 2012. 

\bibitem{Latouche2} J. Wyse, N. Friel and P. Latouche, Inferring structure in bipartite networks using the latent block model and exact ICL,
%\emph{Signal Processing}, 45:59-83, Elsevier, 1995.
\emph{arXiv pre-print} arXiv: 1404.2911, 2014.

\bibitem{Biernacky} C. Biernacky, G. Celeux and G. Govaert, Assessing a mixture model for clustering with the integrated completed likelihood, 
\emph{IEEE Trans. Pattern Anal. Machine Intel}, vol.7, pp. 719-725, 2000.

\bibitem{Latouche1} E. C\^ome and P. Latouche, Model selection and clustering in stochastic block models with the exact integrated complete data likelihood, 
\emph{arXiv pre-print}, arXiv:1303.2962, 2013.

% For articles

\bibitem{Cattuto} L. Isella, J. Sthel\'e, A. Barrat, C.Cattuto, J.F. Pinton, W. Van den Broeck, What's in a crowd? Analysis of face-to-face behavioral networks,
\emph{Journal of Theoretical Biology}, vol. 271, pp. 166-180, 2011.

% For paper in conference proceedings


% For Technical Report
% \bibitem{Stone_TechRep} J. V. Stone and J. Porrill, Undercomplete independent component analysis for signal separation and dimension
% reduction. Technical Report, Psychology Department, Sheffield
% University, Sheffield, S10 2UR, England, October 1997.
\end{thebibliography}
\end{document}